\documentclass[conference]{IEEEtran}
\IEEEoverridecommandlockouts
\usepackage{cite}
\usepackage{amsmath,amssymb,amsfonts}
\usepackage{algorithmic}
\usepackage{graphicx}
\usepackage{textcomp}
\usepackage{xcolor}
\usepackage{booktabs}
\usepackage{multirow}
\def\BibTeX{{\rm B\kern-.05em{\sc i\kern-.025em b}\kern-.08em
    T\kern-.1667em\lower.7ex\hbox{E}\kern-.125emX}}
\begin{document}

\title{TextualVerifier: Verify TextGrad Step-by-Step}

\author{
\IEEEauthorblockN{1\textsuperscript{st} Eugenius Mario Situmorang}
\IEEEauthorblockA{Universitas Indonesia\\
Depok, Indonesia \\
eugenius.mario@ui.ac.id}
\and
\IEEEauthorblockN{2\textsuperscript{nd} Adila Alfa Krisnadhi}
\IEEEauthorblockA{Universitas Indonesia\\
Depok, Indonesia \\
adila@cs.ui.ac.id}
\and
\IEEEauthorblockN{3\textsuperscript{rd} Ari Wibisono}
\IEEEauthorblockA{Universitas Indonesia\\
Depok, Indonesia \\
ari.w@cs.ui.ac.id}
}
\maketitle

\begin{abstract}
TextGrad is a novel approach in text-based automatic differentiation that enables composite AI systems to perform optimization without explicit numerical equations. However, TextGrad currently lacks self-verification mechanisms that are crucial for ensuring reasoning validity in text-based decision making. This research introduces TextualVerifier, a verification framework that leverages chain-of-thought and majority voting with LLMs to address the verification gap in TextGrad optimization. TextualVerifier implements a four-stage workflow: chain-of-thought decomposition, variant generation, majority voting, and consensus aggregation. This framework integrates non-invasively with TextGrad at the loss function stage and optimization result verification. Experimental evaluation is divided into two main phases using the Gemini 1.5 Pro model: (1) standalone evaluation using PRM800K, and (2) integrated evaluation with TextGrad through GPQA-Diamond, MMLU-ML, and MMLU-CP benchmarks. Results demonstrate success rates with statistically significant improvements (p < 0.001). In phase one, TextualVerifier can improve the validity of solution steps by 29\%. In phase two, TextualVerifier achieves highest performance when implemented in TextGrad's loss function with +2.2pp improvement from 68.2\% to 70.4\% with moderate overhead averaging 5.9 LLM calls. Additional evaluation conducted with TextualVerifier versioning for TextGrad loss yields improvements of +8.08pp from 51.01\% to 59.09\% (GPQA), +10.71pp from 76.79\% to 87.50\% (MMLU-ML), and +3.92pp from 91.18\% to 95.10\% (MMLU-CP). TextualVerifier presents the first self-verification approach for TextGrad through LLM-based techniques without requiring numerical gradients, featuring curated verification prompts that enhance reasoning reliability. This framework opens new directions for verification in text-based optimization.
\end{abstract}


\section{Introduction}

The emergence of large language models (LLMs) has revolutionized natural language processing and reasoning tasks, with models like GPT-3 achieving impressive performance across diverse tasks through 175 billion parameters \cite{b1} and demonstrating remarkable capabilities in zero-shot reasoning \cite{b2} and mathematical problem solving \cite{b3}. However, these models remain susceptible to generating factually incorrect information, particularly in complex multi-step reasoning scenarios \cite{b4}.

TextGrad, introduced by Yuksekgonul et al. \cite{b5}, represents a groundbreaking approach to automatic differentiation via text, enabling the optimization of complex objectives through textual gradients and loss functions. This framework demonstrates remarkable effectiveness across diverse applications, improving GPT-4o's zero-shot accuracy in Google-Proof Question Answering from 51\% to 55\%, yielding 20\% relative performance gains in optimizing LeetCode-Hard coding solutions, and achieving 36\% completion rate on complex programming tasks compared to previous 23\% baselines \cite{b5}. Building upon this foundation, recent work has explored meta-learning approaches such as metaTextGrad \cite{b12}, which demonstrates 5-27\% performance gains on question-answering tasks.

However, TextGrad currently lacks self-verification mechanisms, which are crucial for ensuring the reliability and accuracy of generated reasoning processes. The importance of such verification has been demonstrated in recent work showing that large language models can be better reasoners when equipped with verification capabilities \cite{b6}. Recent advances in process supervision have demonstrated significant improvements in training reliable reward models compared to outcome supervision approaches \cite{b7}. Process-supervised reward models (PRMs) provide feedback for each step in the chain-of-thought reasoning, achieving state-of-the-art performance with 78.2\% accuracy on the MATH test set \cite{b7}. This approach has been further validated through deductive verification methods \cite{b13} and stepwise correction techniques \cite{b14}.

The distinction between outcome supervision and process supervision is fundamental to understanding verification methodologies. Outcome-supervised reward models (ORMs) evaluate only the final result of a model's reasoning chain, while process-supervised models provide step-by-step feedback throughout the reasoning process \cite{b8}. This granular approach has proven more effective in detecting and preventing reasoning errors before they compound into larger solution failures, as demonstrated in recent work on faithfulness and factuality in abstractive summarization \cite{b15}.

TextGrad's current architecture lacks integrated verification mechanisms, creating a critical gap in its ability to detect and correct reasoning errors during optimization. Without self-verification capabilities, the system cannot distinguish between correct and incorrect intermediate steps, potentially reinforcing erroneous reasoning patterns through its iterative optimization approach.

This paper introduces TextualVerifier, a self-verification framework using LLM and prompt engineering that addresses the verification gap in TextGrad. The main contributions include: (1) First self-verification framework for TextGrad, filling a critical gap in the framework's capabilities; (2) LLM-based textual verification architecture that leverages large language models without relying on numerical gradients; (3) Textual step-by-step verification methodology establishing a systematic four-stage verification process comprising chain-of-thought decomposition, variant generation, majority voting, and step merging.

\section{Related Work}

\subsection{TextGrad}

TextGrad represents a paradigmatic shift in optimization methodologies for compound AI systems, introducing automatic "differentiation" via text to address the fundamental challenge of optimizing systems orchestrating multiple large language models \cite{b5}. The framework emerges from the recognition that traditional backpropagation cannot be applied to modern AI systems where components communicate through natural language interfaces rather than differentiable mathematical operations.

TextGrad conceptualizes compound AI systems as computation graphs where nodes represent variables (unstructured text) and edges represent dependencies. Unlike traditional numerical computation graphs, TextGrad's variables contain arbitrary textual content including prompts, code snippets, or molecular representations, while maintaining compositional properties essential for optimization. Similarly, DSPy \cite{b16} has shown how declarative language model calls can be compiled into self-improving pipelines, highlighting the growing importance of systematic optimization in LLM applications.

Textual gradients are represented as natural language feedback providing directional guidance for improving variables with respect to downstream objectives. TextGrad defines textual gradients as LLM-generated critiques and suggestions indicating how variables should be modified, leveraging LLMs' semantic understanding to provide contextually appropriate feedback through carefully designed prompts.

\subsection{Zero-Shot Chain-of-Thought Reasoning}

The development of reasoning capabilities in LLMs has progressed through several key paradigms. Kojima et al. \cite{b2} demonstrated that large language models are zero-shot reasoners, showing that simple prompting techniques like "Let's think step by step" can elicit sophisticated reasoning behaviors without providing examples. This finding is crucial for verification systems that must operate without extensive training data.

Building upon zero-shot capabilities, chain-of-thought reasoning has emerged as a fundamental technique for improving LLM reasoning quality. Wang et al. \cite{b9} introduced self-consistency methods that generate multiple reasoning paths and select the most frequent answer, achieving significant performance gains of +17.9\% on GSM8K, +11.0\% on SVAMP, and +12.2\% on AQuA reasoning benchmarks. Tree-of-thought reasoning \cite{b17} further extends this approach by enabling deliberate exploration of reasoning paths through structured problem-solving strategies.

\subsection{Process-Supervised Reward Models}

Two primary paradigms have emerged for training reward models: outcome supervision and process supervision. Uesato et al. \cite{b8} describe these as fundamentally different approaches to providing feedback, with distinct implications for reasoning reliability and error detection.

Process-supervised reward models (PRMs) address ORM limitations by providing feedback for each reasoning step, enabling fine-grained assessment of logical consistency and reasoning quality. Lightman et al. \cite{b7} demonstrate that PRMs receive feedback for each chain-of-thought step, allowing early detection and correction of reasoning errors before propagation. PRMs achieve state-of-the-art performance with 78.2\% accuracy on the MATH test set, significantly outperforming outcome-only approaches. The MATH dataset \cite{b18} provides a comprehensive benchmark for measuring mathematical problem-solving capabilities across competition-level mathematics problems.

\subsection{Self-Verification and Correction}

Self-verification in large language models represents a paradigm where models evaluate their own reasoning processes and outputs, enabling autonomous error detection and correction without external validation systems. Weng et al. \cite{b6} demonstrate that large language models can be better reasoners when equipped with self-verification capabilities, showing significant improvements in reasoning accuracy through internal validation mechanisms.

Recent advances include revision after explanation \cite{b10} and self-verification with self-correction for improved test-time scaling \cite{b11}. These approaches demonstrate the potential for language models to improve their reasoning through iterative refinement processes, providing a foundation for integrating verification directly into optimization loops. Contemporary research has demonstrated the effectiveness of various verification approaches, including deductive verification of chain-of-thought reasoning \cite{b13} and enhancing mathematical reasoning through stepwise correction methodologies \cite{b14}.

\section{TextualVerifier}

\subsection{Architecture Design}

TextualVerifier is designed as a modular verification framework that integrates seamlessly with TextGrad's existing architecture while providing comprehensive step-by-step verification capabilities. The architecture leverages the principle of process supervision to enable fine-grained assessment of reasoning chains and systematic error detection during textual optimization.

\begin{figure*}[htbp]
\centerline{\includegraphics[width=\textwidth]{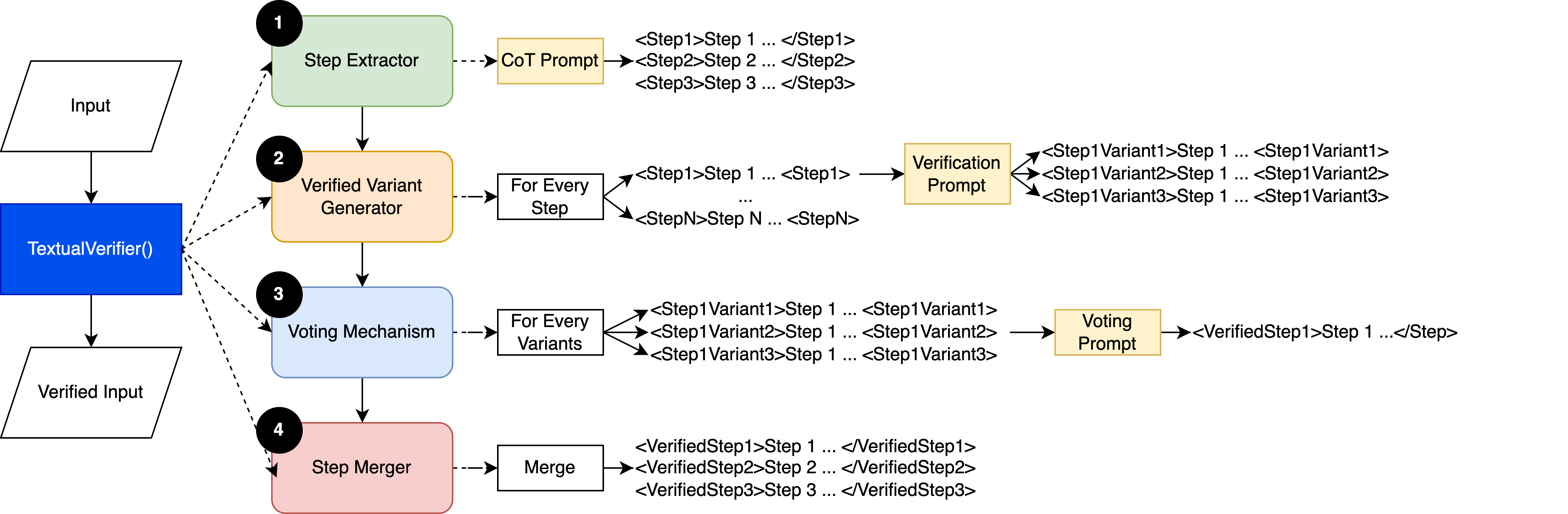}}
\caption{TextualVerifier Architecture showing the four-stage verification workflow: (1) Chain-of-Thought Decomposition, (2) Step Breakdown and Extraction, (3) Variant Generation with Multiple Perspectives, and (4) Majority Voting and Consensus Aggregation.}
\label{fig:architecture}
\end{figure*}

The TextualVerifier architecture consists of four primary components that work together to provide comprehensive verification capabilities:

\begin{enumerate}
\item \textbf{Step Extractor}: Decomposes complex reasoning chains into individual logical steps using chain-of-thought prompting techniques. This component extracts steps using regex patterns for \texttt{<STEP>...</STEP>} tags and provides fallback mechanisms for unstructured text.

\item \textbf{Verified Variant Generator}: Creates multiple alternative formulations of each reasoning step to capture different verification perspectives and potential interpretations of the logical flow. This component applies multiple verification task prompts to generate independent assessments.

\item \textbf{Voting Mechanism}: Evaluates and selects the best variant through majority voting consensus to ensure robust decision-making. This component aggregates verification results using majority voting to identify the most frequently occurring pattern among variants.

\item \textbf{Step Merger}: Combines verified steps into final output with standardized formatting using \texttt{<VERIFIED>} tags, producing outputs like: \texttt{<VERIFIED>Step 1: Calculate discriminant</VERIFIED><VERIFIED>Step 2: Apply quadratic formula</VERIFIED>}.
\end{enumerate}

The verification process implements a systematic four-stage verification process:

\textbf{Stage 1: Chain-of-Thought Decomposition} - The first stage processes input reasoning chains to generate explicit step-by-step reasoning when required through structured prompting.

\textbf{Stage 2: Step Breakdown and Extraction} - The second stage extracts individual reasoning steps from the structured output, creating a list of discrete logical operations for verification using both explicitly tagged steps and fallback mechanisms for unstructured text.

\textbf{Stage 3: Variant Generation} - The third stage creates multiple alternative formulations of each reasoning step using different verification perspectives, addressing the inherent variability in LLM outputs by generating multiple independent assessments.

\textbf{Stage 4: Majority Voting and Consensus} - The final stage aggregates verification results using majority voting to select the best variant for each reasoning step, reducing the impact of individual verification errors while preserving high-quality reasoning.

\subsection{TextGrad Integration}

The integration of TextualVerifier with TextGrad requires careful consideration of the optimization workflow to ensure verification enhances rather than disrupts the optimization process. The integration strategy implements verification at two critical points in the TextGrad optimization cycle: during loss calculation evaluation and after optimization result generation.

\begin{figure}[htbp]
\centerline{\includegraphics[width=\columnwidth]{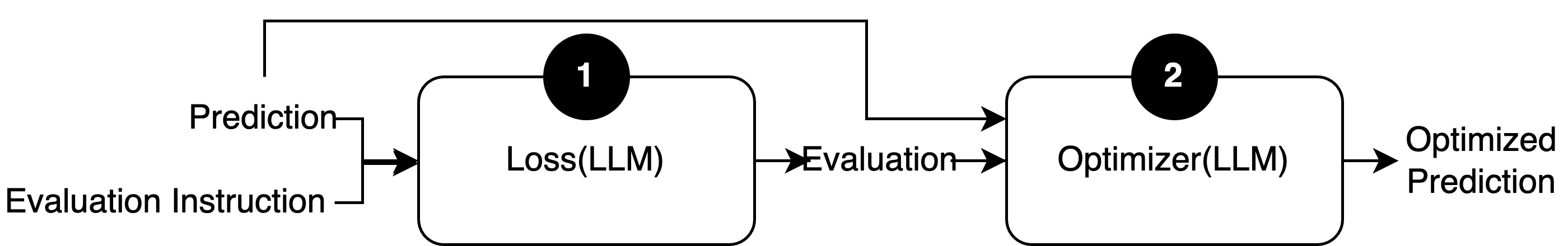}}
\caption{Integration points with TextGrad showing Loss Function Verification and Optimization Phase Verification within the TextGrad optimization workflow.}
\label{fig:integration_points}
\end{figure}

\textbf{Loss Function Verification} addresses the fundamental challenge of ensuring that loss calculations accurately reflect solution quality. The loss verification phase implements the verification formula:

\begin{align}
&\text{instance} + \text{instruction} \Rightarrow \text{loss\_value} \\
&\text{instance} + \text{loss\_value} + \text{verification\_prompt} \nonumber \\
&\qquad \Rightarrow \text{verified\_loss\_value}
\end{align}

Loss function verification integrates with TextGrad's TextLoss component through a wrapper approach that preserves the original loss calculation interface while adding verification capabilities.

\textbf{Optimization Phase Verification} validates the output generated by TextGrad's optimization process, ensuring that optimized solutions maintain logical correctness while addressing the issues identified in the loss function. This verification phase implements the formula:

\begin{align}
&(\text{initial\_solution} \land \text{loss\_value}) + \text{optimization\_instruction} \nonumber \\
&\qquad \Rightarrow \text{optimized\_solution} \\
&(\text{initial\_solution} \land \text{loss\_value}) + \text{optimized\_solution} \nonumber \\
&\qquad + \text{verification\_prompt} \Rightarrow \text{verified\_optimized\_solution}
\end{align}

The optimization verification integrates with TextGrad's TextualGradientDescent optimizer through a post-optimization verification step that systematically evaluates each step for correctness and applies targeted corrections where errors are detected.

\section{Experiment}

\subsection{Configuration}

The experimental configuration is designed to ensure reproducibility and systematic evaluation across all test conditions. The evaluation employs Gemini 1.5 Pro as the base LLM engine with 2M token context window capability, which enables comprehensive verification of complex reasoning chains.

The experiment is divided into two main sessions: (1) TextualVerifier standalone evaluation using PRM800K dataset containing 1,315 steps from 70 questions, and (2) TextGrad with TextualVerifier integration using GPQA-Diamond (198 questions), MMLU Machine Learning (112 questions), and MMLU College Physics (102 questions) datasets.

For session 1, evaluation employs Stuart-Maxwell Test for rating improvement analysis and answer accuracy assessment. For session 2, evaluation uses McNemar's Test for accuracy improvement measurement across three integration configurations: loss verification only, optimizer verification only, and combined verification.

\subsection{TextualVerifier Only}

The first experimental phase evaluated TextualVerifier's standalone performance across five different configurations using the preprocessed PRM800K dataset. Each configuration employed varying numbers of verification perspectives (1-5 variants) to assess multi-perspective consensus effectiveness and computational trade-offs.

\begin{figure}[htbp]
\centerline{\includegraphics[width=\columnwidth]{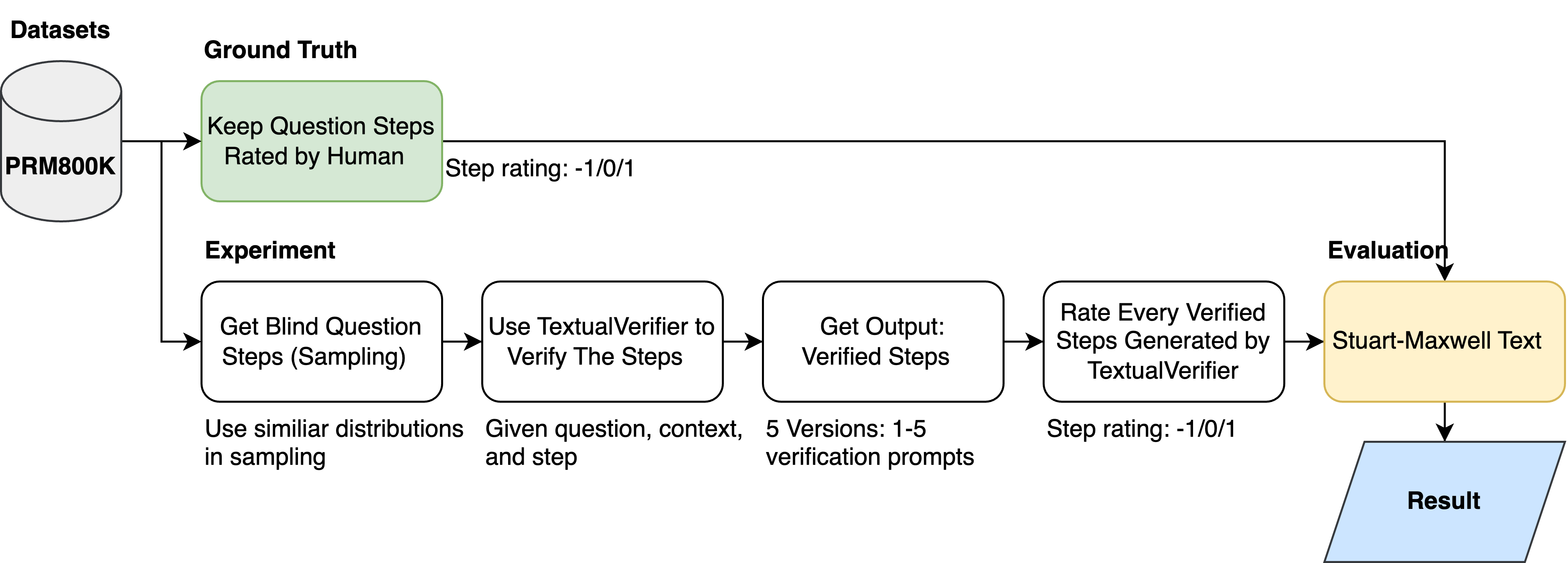}}
\caption{Experiment Phase 1 High-Level Flow using PRM800K dataset.}
\label{fig:exp_phase1}
\end{figure}

The PRM800K dataset required extensive preprocessing due to structural complexity. The raw dataset contained deeply nested JSON structures with multiple completion alternatives for each reasoning step. A sampling algorithm was developed to create deterministic reasoning paths while preserving natural error distribution patterns.

The evaluation processed 1,315 reasoning steps from 70 mathematical problems, with systematic performance tracking including processing time, LLM calls, token consumption, and verification stages.

\subsection{TextGrad with TextualVerifier}

The second experimental phase evaluated TextualVerifier integration within the TextGrad optimization framework across four distinct configurations and three academic datasets. The evaluation processed 412 questions to assess verification impact on optimization effectiveness.

\begin{figure}[htbp]
\centerline{\includegraphics[width=\columnwidth]{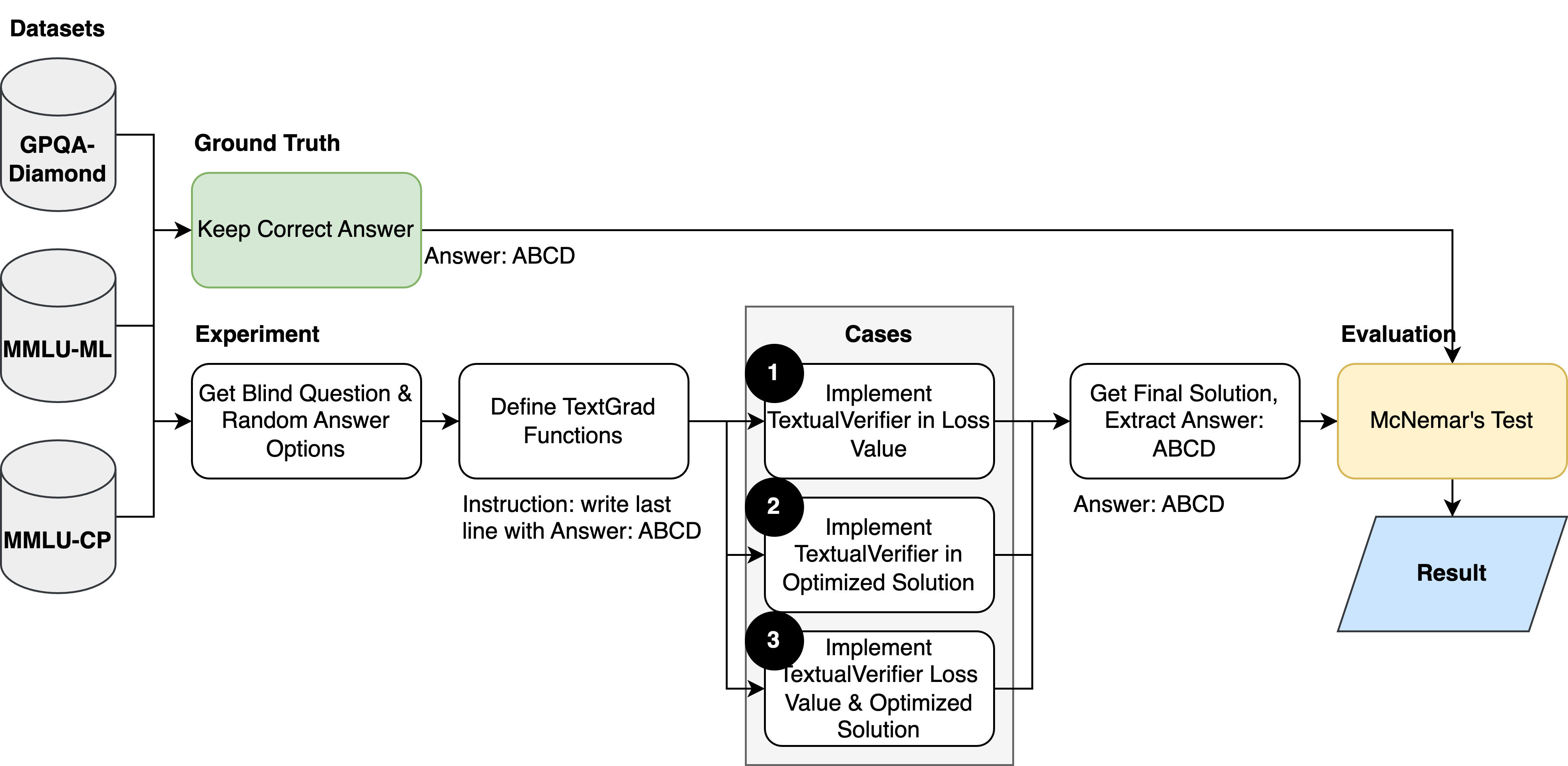}}
\caption{Experiment Phase 2 High-Level Flow using GPQA-Diamond, MMLU-ML, and MMLU-CP datasets.}
\label{fig:exp_phase2}
\end{figure}

Initial solution generation employed zero-shot prompting with systematic answer randomization to ensure unbiased evaluation. The process generated reasoning chains that serve as starting points for TextGrad optimization.

Four distinct configurations were systematically evaluated:
\begin{enumerate}
\item \textbf{TextGrad Only}: Baseline performance without verification
\item \textbf{TextGrad + TV (Loss)}: Loss value verification only
\item \textbf{TextGrad + TV (Optimizer)}: Optimization result verification only  
\item \textbf{TextGrad + TV (Both)}: Combined verification at both stages
\end{enumerate}

\subsection{Additional Evaluation in Loss}

Building upon the comprehensive evaluation framework, an additional evaluation phase implemented systematic comparative assessment of TextualVerifier versions V1-V4. This evaluation focused exclusively on loss function verification to isolate architectural and prompt engineering effects.

The four TextualVerifier versions implement distinct architectural approaches:
\begin{itemize}
\item \textbf{V1}: Sequential individual step processing with five-stage pipeline
\item \textbf{V2}: Context-enhanced sequential processing with cumulative context integration
\item \textbf{V3}: Consolidated parallel processing achieving ~70\% cost savings
\item \textbf{V4}: Hybrid simplification approach with dual-layer verification
\end{itemize}

The evaluation implemented controlled experimental design focusing exclusively on loss function verification to isolate architectural effects across the four TextualVerifier versions using identical datasets and evaluation metrics.

\section{Results and Discussion}

\subsection{TextualVerifier Only}

The first experimental phase revealed significant differences in accuracy improvement, processing efficiency, and computational overhead across the different multi-perspective configurations. All five TextualVerifier configurations achieved 100\% success rates in processing the complete dataset.

Version 1 achieved optimal balance between accuracy improvement and computational efficiency (+1.4 pp improvement, 66.5 seconds processing time), while Version 4 delivered the highest accuracy improvement (+5.7 pp) at moderate computational cost. Version 5 exhibited unexpected performance degradation with zero accuracy improvement despite maximum resource consumption.

The Stuart-Maxwell test analysis ($\alpha = 0.05$, $df = 2$) provided rigorous statistical validation with all configurations demonstrating highly significant evidence against marginal homogeneity ($p < 0.001$). The rating transition analysis revealed systematic positive bias with 28.8-29.0\% of steps showing rating improvements, while only 0.9-1.3\% showed degradations.

\begin{table}[htbp]
\caption{TextualVerifier Standalone Performance}
\begin{center}
\begin{tabular}{|c|c|c|c|c|}
\hline
\textbf{Config} & \textbf{Accuracy} & \textbf{Proc. Time} & \textbf{LLM Calls} & \textbf{Score} \\
\hline
1 variant & +1.4 pp & 66,465 ms & 18.8 & 66.3/100 \\
\hline
2 variants & +1.4 pp & 196,098 ms & 56.4 & 63.1/100 \\
\hline
3 variants & +2.9 pp & 227,156 ms & 75.1 & 62.2/100 \\
\hline
4 variants & +5.7 pp & 253,620 ms & 93.9 & 57.9/100 \\
\hline
5 variants & +0.0 pp & 327,053 ms & 112.7 & 32.9/100 \\
\hline
\end{tabular}
\label{tab1}
\end{center}
\end{table}

\subsection{TextGrad with TextualVerifier}

The comprehensive evaluation across three datasets revealed critical insights for optimal TextualVerifier deployment, with distinct patterns emerging based on domain characteristics and question complexity.

\textbf{Overall Performance}: TextGrad + TV (Loss) achieved the highest overall accuracy (70.4\%) with moderate computational overhead, representing a +2.2 percentage point improvement over baseline TextGrad (68.2\%). This configuration demonstrates optimal integration effectiveness with an average of 5.9 LLM calls and 11,944 ms processing time.

\textbf{Dataset-Specific Patterns}:
\begin{itemize}
\item \textbf{GPQA-Diamond}: Combined approach achieved highest accuracy (57.1\%, +4.1 pp), suggesting complex graduate-level reasoning benefits from comprehensive verification
\item \textbf{MMLU Machine Learning}: Loss-only verification showed strongest response (+2.7 pp to 80.4\%), while optimizer verification caused significant degradation (-18.8 pp)
\item \textbf{MMLU College Physics}: Robust improvements across all configurations (+1.9 to +4.9 pp), with combined approach achieving highest accuracy (92.2\%)
\end{itemize}

\begin{table}[htbp]
\caption{TextGrad with TextualVerifier Integration Performance}
\begin{center}
\begin{tabular}{|l|c|c|c|}
\hline
\textbf{Method} & \textbf{Accuracy} & \textbf{Improvement} & \textbf{LLM Calls} \\
\hline
TextGrad-Only & 68.2\% & - & 0.0 \\
\hline
TextGrad + TV (Loss) & \textbf{70.4\%} & \textbf{+2.2 pp} & 5.9 \\
\hline
TextGrad + TV (Optimizer) & 65.0\% & -3.2 pp & 9.6 \\
\hline
TextGrad + TV (Both) & 67.2\% & -1.0 pp & 15.1 \\
\hline
\end{tabular}
\label{tab2}
\end{center}
\end{table}

\subsection{Additional Evaluation in Loss}

The version comparison evaluation revealed critical insights about architectural effectiveness across different academic domains. V2 and V3 demonstrated joint leadership with both achieving identical zero-shot accuracy (73.30\%) and improvement (+5.34 percentage points) over baseline.

\textbf{Domain-Specific Version Effectiveness}:
\begin{itemize}
\item \textbf{GPQA-Diamond}: V3 achieved highest zero-shot improvement (+8.08 pp), demonstrating exceptional capability for complex multi-disciplinary scientific problems
\item \textbf{MMLU Machine Learning}: V3 showed exceptional zero-shot improvement (+10.71 pp to 87.50\%), representing the highest single improvement across all datasets
\item \textbf{MMLU College Physics}: V1 achieved highest initial improvement (+3.92 pp), suggesting systematic step-by-step verification provides immediate benefits for physics reasoning
\end{itemize}

\begin{table}[htbp]
\caption{TextualVerifier Version Comparison Results}
\begin{center}
\begin{tabular}{|l|c|c|c|c|}
\hline
\textbf{Version} & \textbf{GPQA} & \textbf{MMLU-ML} & \textbf{MMLU-CP} & \textbf{Overall} \\
\hline
TextGrad Only & 51.01\% & 76.79\% & 91.18\% & 67.96\% \\
\hline
V1 (Basic) & +5.05 pp & -2.68 pp & +3.92 pp & +2.67 pp \\
\hline
V2 (Contextual) & +5.56 pp & +7.14 pp & +2.94 pp & \textbf{+5.34 pp} \\
\hline
V3 (Consolidated) & \textbf{+8.08 pp} & \textbf{+10.71 pp} & +0.98 pp & \textbf{+5.34 pp} \\
\hline
V4 (Simplified) & +1.01 pp & +5.35 pp & -2.94 pp & +1.21 pp \\
\hline
\end{tabular}
\label{tab3}
\end{center}
\end{table}

\subsection{Dataset Issues}

The research encountered significant challenges with the PRM800K dataset that substantially impacted the experimental design. Analysis revealed that only 43.5\% of reasoning chains reached successful completion, with 56.5\% terminating due to complexity barriers or systematic errors.

The preprocessing pipeline resulted in dramatic data reduction from 2,868 to 70 questions (97.6\% reduction), highlighting critical dataset quality issues. The multi-completion structure created exponential path combinations (256 to 65,536 possible chains per question) requiring specialized sampling algorithms.

\subsection{Ablation Study}

The ablation study systematically evaluated the impact of TextualVerifier integration at different stages, revealing stage-specific effects on performance and domain sensitivity.

\textbf{Loss Value Verification Impact}: Demonstrated consistent positive effects across all evaluated datasets, establishing it as the most reliable integration approach. Achieved +2.1 to +2.7 pp improvements with moderate LLM usage (4.3-6.9 calls per problem).

\textbf{Optimizer Result Verification Impact}: Exhibited variable effects with significant domain-specific sensitivity. Showed +2.1 to +3.9 pp improvements in GPQA-Diamond and MMLU-CP, but -18.8 pp degradation in MMLU-ML, indicating potential interference with domain-specific optimization patterns.

\textbf{Combined Verification Analysis}: Achieved moderate improvements with significant computational overhead. Demonstrated complex interactions between verification stages, with additive benefits in some domains but insufficient overall value proposition compared to loss-only verification.

\section{Conclusion}

This research successfully addresses the critical verification gap in TextGrad optimization through the introduction of TextualVerifier, a comprehensive self-verification framework leveraging chain-of-thought reasoning and majority voting mechanisms. The experimental results demonstrate statistically significant improvements across multiple evaluation dimensions while establishing optimal deployment strategies for different academic domains.

Key findings include: (1) TextualVerifier achieves 29\% improvement in reasoning step validity through systematic process supervision, (2) Loss function verification provides optimal integration approach with +2.2 pp overall improvement and moderate computational overhead, (3) Version-specific evaluation reveals V2 and V3 as optimal architectures with +5.34 pp zero-shot improvements across diverse domains.

The research establishes TextualVerifier as the first self-verification framework for TextGrad, demonstrating that LLM-based verification can enhance reasoning reliability without requiring numerical gradients. The modular architecture enables flexible deployment across different optimization scenarios while maintaining computational efficiency.

Future work should focus on expanding verification capabilities to multimodal contexts, developing domain-adaptive verification strategies, and investigating integration with other textual optimization frameworks. The demonstrated effectiveness of process-supervised verification opens new directions for reliable AI system development.

\section*{Acknowledgment}

The authors thank the Information Retrieval \& Natural Language Processing Lab at Universitas Indonesia for providing computational resources and research support. We acknowledge the contributions of the open-source community for providing access to datasets and evaluation frameworks that enabled this research.

\end{document}